\documentclass[letterpaper]{article} 
\usepackage{aaai2027}  
\usepackage[hyphens]{url}  
\usepackage{graphicx} 
\usepackage{natbib}  
\usepackage{caption} 
\usepackage{xcolor}
\usepackage{soul}
\soulregister\sys0
\soulregister\cite7

\definecolor{colorE}{rgb}{0.9, 0.95, 1.0} 
\definecolor{colorRatio}{rgb}{0.95, 0.95, 0.95} 
\definecolor{highlightcolor}{rgb}{0.92, 0.96, 1.0}
\usepackage{amsmath,amssymb}
\usepackage{comment}
\usepackage{pifont}
\usepackage{booktabs}
\usepackage{multirow}
\usepackage{colortbl}
\usepackage{tabularx}
\usepackage{subcaption}
\usepackage{makecell}
\usepackage{algorithm}
\usepackage{algorithmic}

\newcommand{\sys}{E-AdaPrune}

\usepackage{enumitem}
\usepackage{newfloat}
\usepackage{listings}
\DeclareCaptionStyle{ruled}{labelfont=normalfont,labelsep=colon,strut=off} 
\floatstyle{ruled}
\newfloat{listing}{tb}{lst}{}
\floatname{listing}{Listing}

\usepackage{booktabs}

\nocopyright

\title{Energy-Driven Adaptive Visual Token Pruning for \\ Efficient Vision-Language Models}
\author{
    Jialuo He, Huangxun Chen\corresponding
}
\affiliations{
    The Hong Kong University of Science and Technology (Guangzhou)\\
    huangxunchen@hkust-gz.edu.cn

}

\begin{document}

\maketitle

\begin{abstract}
  Visual token reduction is critical for accelerating Vision-Language Models (VLMs), since visual inputs are represented as token sequences that introduce substantial computational overhead in the LLM backbone. However, most pruning pipelines treat efficiency primarily as a token selection problem and retain a fixed visual token budget across inputs, overlooking the substantial variation in image information density. We propose \sys{}, an energy driven adaptive pruning framework that determines an image specific token budget from the singular value spectrum of the visual feature matrix and passes this budget to existing token selectors. By preserving a certain proportion of spectral energy, our method allocates more tokens to information dense scenes while assigning fewer tokens to redundant scenes, without introducing additional learnable parameters. We evaluate \sys{} across four VLM backbones, three token selectors, and nine benchmarks under matched average token budgets. 
  Results show that \sys{} removes a substantial amount of redundant computation from simple cases and converts the saved budget into larger gains on information rich cases.
  Notably, on SQA$^\mathrm{I}$ with Qwen2.5-VL-3B, \sys{} uses 35.8\% fewer tokens for simple cases with only a 0.52\% relative performance decrease. The saved budget is redirected to hard cases, which receive 52.5\% more tokens and achieve a 1.94\% relative performance improvement. 
\end{abstract}


\section{Introduction}
\label{sec:intro}
Large Vision-Language Models (LVLMs) \cite{blip2,gpt4v,qwen2-vl,Qwen-VL,glm,improvedllava,llava,llavanext,minigpt} have demonstrated exceptional capabilities across diverse multimodal tasks, including complex visual reasoning, fine-grained instruction following, and document understanding \cite{pope,seed,textvqa}. To capture the necessary semantic depth, these models represent visual inputs as sequences of visual tokens. These visual tokens are then consumed by the Large Language Model (LLM) backbone, where longer visual sequences introduce significant computational overhead due to the quadratic complexity of self attention \cite{attention, survey1,survey2,survey3}. Empirical evidence suggests that visual tokens inherently contain substantial redundancy, with only a small fraction of tokens being essential for accurate response generation \cite{3m,blip2,fastv,vocollama}.

\begin{figure}[t]
    \centering
    \includegraphics[width=\linewidth]{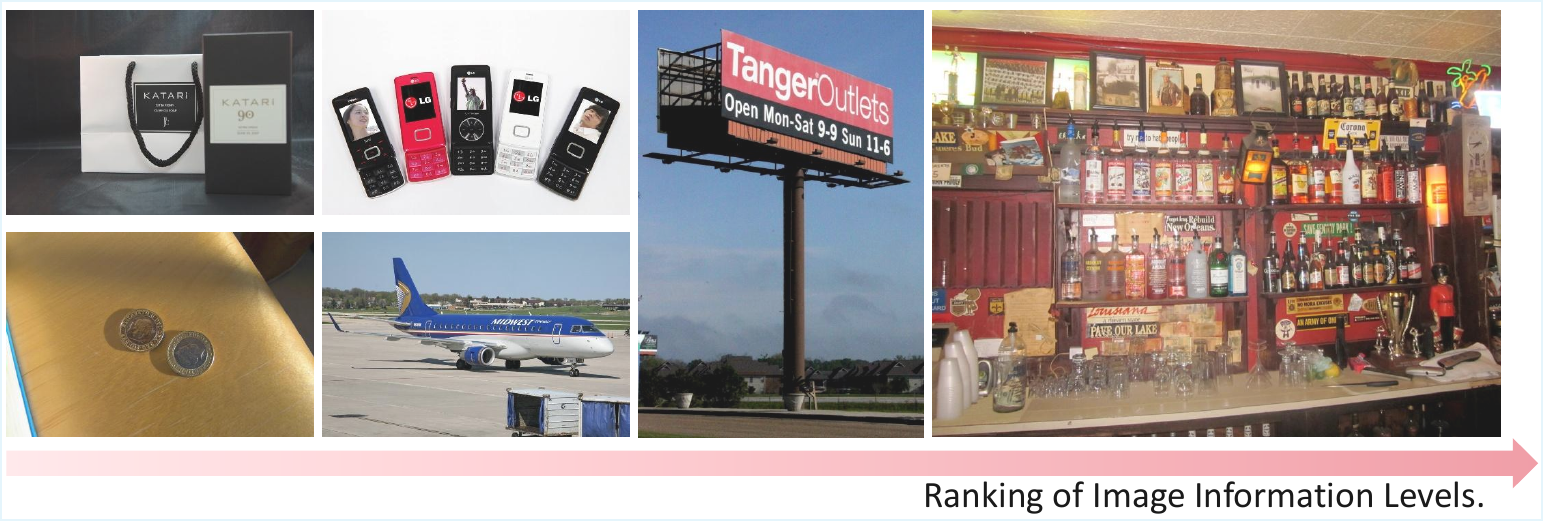}
    \caption{Different images contain different amounts of visual information, indicating that a static token budget may either discard critical details or retain unnecessary redundancy. All examples are from TextVQA \cite{textvqa}.} \vspace{-0.1in}
    \label{fig:energyimgs}
\end{figure}

Existing visual token reduction methods typically shorten the visual sequence by first specifying a retained budget, such as a fixed top-$k$ or a fixed pruning ratio, and then selecting or merging tokens according to attention, similarity, or visual saliency scores \cite{fastv,pdrop,visionzip}. While this design improves how visual tokens are scored, the retained budget itself is often fixed before observing the input image. Such static approaches fail to account for the significant variability in information density between different images. For example, as shown in Figure~\ref{fig:energyimgs}, in an information dense scene like a crowded bar containing numerous legible labels, a model might require more tokens to correctly identify a specific brand, whereas a simple scene with a few mobile phones can be accurately processed with fewer tokens. Applying a single input agnostic budget can over prune complex scenes, causing information loss, while under pruning simple scenes and wasting computation.

This suggests that visual token reduction should decide both which tokens to retain and how many tokens should remain for each image. We therefore view \emph{visual token budgeting as a problem orthogonal to token selection}: a budgeter estimates the retained length, while a selector determines the retained token identities. Recent adaptive methods~\cite{atpllava,vcm,visionthink} attempt to vary the visual length, yet they often rely on extra cost on training, optimization, or policy learning to enable dynamic behavior. We wonder \emph{whether a training-free signal from the visual representation itself can provide an input-specific budget}.

Singular value analysis provides a natural route to such a signal. Prior work has used singular spectra to summarize image signals and feature matrices, showing that spectral concentration captures how information is distributed across dominant components \cite{svd1,svd2,svd3,svd4,svd5}. In VLMs, the visual tokens produced by the vision encoder form a visual feature matrix. When this matrix has a steep spectral decay, most energy is concentrated in a few dominant directions, indicating stronger redundancy and a smaller required token budget. When the spectrum is flatter, the representation spreads energy across more directions, suggesting higher information density and a larger required token budget.

Inspired by the above observations, we propose \sys{}, a training-free and plug-and-play framework for adaptively determining image-specific token budgets. \sys{} converts the singular spectrum of the visual feature matrix into an image-specific token budget by preserving a target fraction $\tau$ of the total spectral energy. 
Compared with directly specifying the token budget, the energy ratio $\tau$ enables the budget to be adaptively determined for each image according to its information richness, while providing a smooth operating point for balancing accuracy and token compression. A larger value preserves more spectral components and consequently retains more visual tokens, whereas a smaller value promotes more aggressive compression. 
Because \sys{} estimates only the number of tokens to retain, it can be seamlessly integrated with a broad range of token selectors~\cite{fastv,pdrop,visionzip} without modifying their token scoring mechanisms. 
Our contributions are summarized as follows:
\begin{itemize}[leftmargin=*]
    \item We identify the practical limitations of fixed token budgets in existing visual-token pruning methods for VLMs and propose leveraging a training-free signal derived directly from visual representations to determine an adaptive and input-specific token budget without incurring the substantial overhead of learnable approaches. 
    \item We decouple visual compression into two stages: image-specific token-budget determination and subsequent top-$k$ token selection. 
    From an energy-based perspective, \sys{} infers an adaptive token budget from the spectral energy of visual features and seamlessly integrates it with a broad range of token-selection strategies. 
    \item We validate \sys{} across 4 VLM backbones, 3 token selectors, and 9 benchmarks. Under matched average budgets, \sys{} removes a substantial amount of redundant computation from simple cases and converts the saved budget into gains on information rich cases. Notably, on SQA$^\mathrm{I}$ with Qwen2.5-VL-3B, \sys{} uses 35.8\% fewer tokens for simple cases with only a 0.52\% relative performance decrease. The saved budget is redirected to hard cases, which receive 52.5\% more tokens and achieve a 1.94\% relative performance improvement. 
    \sys{} incurs minimal runtime overhead when integrated with VLM visual-token pruning methods. With randomized singular value decomposition, the extra overhead is only 8ms per image. 
\end{itemize}

\section{Related Work}
\label{sec:relatedwork}
\noindent\textbf{Visual Token Redundancy in VLMs.}
Empirical studies have demonstrated that only a fraction of visual tokens are essential for VLMs to generate accurate responses~\cite{atpllava,holov,ocr1,sparsevlm,SparseVILA,pdrop,fastv,vcm,vocollama,vispruner,visionzip,visionthink,vtw,3m}. 
Redundant visual tokens impose substantial computational and memory overhead, primarily owing to the quadratic complexity of self-attention in LLM backbone~\cite{survey1, survey2}.

\noindent\textbf{Fixed-Budget Visual Token Reduction.}
Existing visual token compression methods can be broadly categorized into feature abstraction, token merging, and token dropping. Feature abstraction methods resample visual features into a fixed number of latent tokens using learnable query-based bottlenecks, producing constant-length representations independent of input resolution, as exemplified by BLIP-2 \cite{blip2}, InstructBLIP \cite{instructblip}, and MiniGPT-4 \cite{gpt4}. 
Token merging methods reduce token cardinality by aggregating similar patches into representative embeddings based on similarity metrics or hierarchical grouping \cite{tome,prunemerge,ficoco,llavolta,vflowopt}.
Token dropping methods instead discard uninformative tokens to accelerate inference. 
Prior work estimates token importance using cross-modal or self-attention scores \cite{fastv,sparsevlm,madtp} or pure vision signals \cite{visionzip,vispruner}, and progressively increases sparsity across layers~\cite{pdrop}. 
Although these approaches differ in scoring mechanisms and pruning locations, most of them adopt a fixed top-$k$ or predefined pruning ratio shared across all inputs, thereby ignoring the variability of information density across images.
In contrast, \sys{} focuses on token budgeting: it estimates an image-specific $k^*$ from spectral energy and then hands this budget to existing token selectors.

\noindent\textbf{Adaptive Visual Token Reduction.}
To overcome the limitations of fixed budgets, recent work has explored adaptive token reduction mechanisms. 
ATP-LLaVA \cite{atpllava} introduces learnable thresholds to determine instance-specific and layer-wise pruning ratios through an adaptive token pruning module. 
VCM \cite{vcm} formulates adaptive compression as a vision concept modeling problem, dynamically extracting task-relevant concepts via a forward–backward optimization procedure. 
VisionThink \cite{visionthink} further employs reinforcement learning to decide whether low-resolution inputs suffice or higher-resolution images are necessary for reasoning. 
These methods pay substantial costs associated with training, optimization, or policy learning to enable dynamic behavior. 
In contrast, \sys{} computes an image-specific budget directly from the visual features already produced during inference, without threshold learning, iterative optimization, or policy training. Because budget determination is decoupled from token scoring, the same spectral budgeter can be integrated with selectors based on cross-modal attention, progressive pruning, or image-only redundancy.

\section{Method}
\label{sec:method}
\subsection{Preliminaries of Vision-Language Models}
Modern VLMs typically consist of three core components: a vision encoder $\mathcal{E}_V$, a projector $\mathcal{P}$, and an LLM $\mathcal{L}$ \cite{blip2,flamingo,llava,improvedllava,llavanext}. $\mathcal{E}_V$ maps an input image $\mathcal{X}^V$ into a visual feature space:
\begin{equation}
\mathbf{Z}^V = \mathcal{E}_V(\mathcal{X}^V) \in \mathbb{R}^{n_v \times d_v},
\end{equation}
where $n_v$ and $d_v$ denote the number of output visual tokens and their feature dimensions. Subsequently, the projector $\mathcal{P}$ aligns these visual tokens into the text embedding space:
\begin{equation}
\mathbf{H}^V = \mathcal{P}(\mathbf{Z}^V) \in \mathbb{R}^{n_v \times d_t},
\end{equation}
where $d_t$ is the text hidden dimension. The LLM, composed of a text encoder $\mathcal{E}_t$ and an $M$-layer Transformer decoder $F$, processes the concatenated visual and textual features:
\begin{equation}
\mathbf{Y} = F\big[\mathbf{H}^V; \mathcal{E}_t(\mathcal{X}^T)\big].
\end{equation}
Here, $\mathcal{X}^T$ represents the input text, including system prompts, user queries, and dialogue history, while $[\cdot;\cdot]$ denotes the concatenation operation and $\mathbf{Y}$ is the generated output sequence. 
Since the computational complexity of the self-attention mechanism grows quadratically with the sequence length, the number of visual tokens $n_{v}$ remains a primary efficiency bottleneck of VLMs.

\subsection{\sys{}: Energy-Based Adaptive Pruning}

\subsubsection{Idea Overview.}
\begin{figure}[t]
    \centering
    \includegraphics[width=\linewidth]{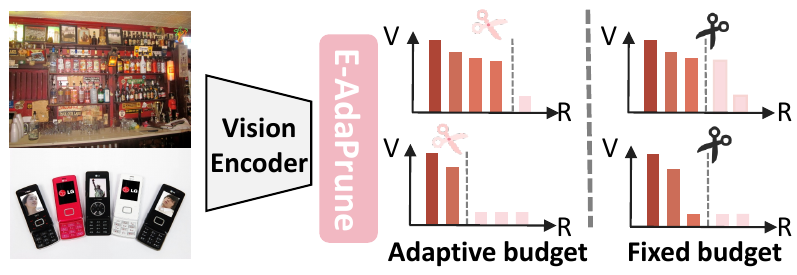}
    \caption{Comparison of static and adaptive pruning. \sys{} determines a content-aware budget $k^*$ via an image-specific energy criterion, optimizing token retention for varying information densities. $V$ and $R$ denote token importance scores and rankings.}\vspace{-0.1in}
    \label{fig:e-adaprune}
\end{figure}

The core idea of \sys{} is to replace the fixed token budget \(k\) with an image-specific budget inferred from the spectral energy of visual features. As illustrated in Fig. \ref{fig:e-adaprune}, static pruning imposes the same top-\(k\) budget on every input, whereas \sys{} first estimates an image-specific token budget \(k^*\) and subsequently passes it to the downstream token selector. Although document understanding and high-precision Optical Character Recognition (OCR) often demand a larger or even complete token budget \cite{ocr1}, we argue that such requirements can be identified directly from the image feature space, obviating the need for task-specific heuristics. 

To implement this observation, we decouple visual compression into two stages: an \emph{energy-based image-specific token budget $k$ determination} stage followed by a \emph{top-$k$ token selection} stage. This decoupling offers substantial architectural flexibility, enabling the dynamic and image-specific budgets $k$ inferred by \sys{} to be seamlessly combined with a broad range of top-\(k\) token selection strategies. These strategies have been actively investigated in recent years and include both text-aware selectors, such as FastV \cite{fastv} and PyramidDrop \cite{pdrop}, and image-only selectors, such as VisionZip \cite{visionzip}. 
In essence, \sys{} provides a plug-and-play solution for determining \emph{how many visual tokens} should be retained, while allowing stronger selectors developed in the future to determine \emph{which visual tokens} to preserve, thereby enabling further performance improvements.

\begin{algorithm}[tb]
\caption{Image-specific Token Budget Determination}
\label{alg:energy_rank}
\textbf{Input}: Visual Features $\mathbf{Z}^V \in \mathbb{R}^{n_v \times d_v}$, Energy Ratio $\tau$ \\
\textbf{Output}: Token Budget $k^*$
\begin{algorithmic}[1]
\STATE $(\mathbf{U}, \mathbf{S}, \mathbf{V}^\top) \leftarrow \text{SVD}(\mathbf{Z}^V)$
\STATE $\sigma \leftarrow \text{diag}(\mathbf{S})$
\STATE $n=\min(n_v,d_v)$
\STATE $\text{E}_{\text{total}} \leftarrow \sum_{i=1}^{n} \sigma_i^2$

\FOR{$k = 1$ to $n$}
    \STATE $C(k) \leftarrow \frac{\sum_{i=1}^k \sigma_i^2}{\text{E}_{\text{total}}}$
    \IF{$C(k) \ge \tau$}
        \STATE $k^* \leftarrow k$
        \STATE \textbf{break}
    \ENDIF
\ENDFOR
\STATE \textbf{return} $k^*$
\end{algorithmic}
\end{algorithm}

\begin{figure*}[t]
    \centering
    \includegraphics[width=1\linewidth]{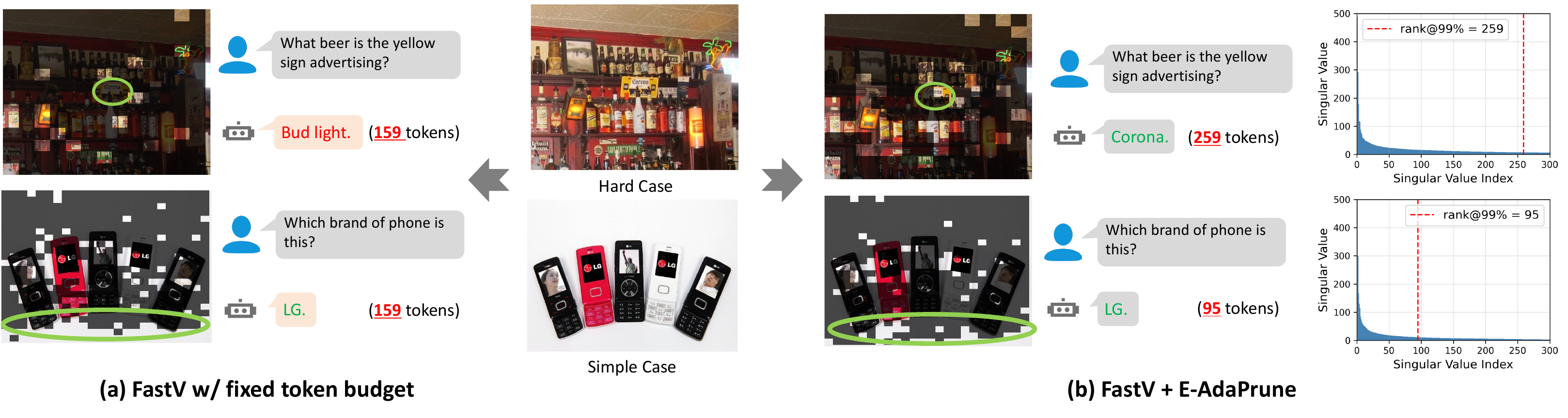}\vspace{-0.1in}
    \caption{Representative examples illustrating the benefits of controlling the visual token budget from an energy perspective. The hard case contains substantially richer information than the simple case. (a) Under a fixed token budget, FastV uniformly retains 159 tokens for both cases. This budget is redundant for the simple case but insufficient to preserve critical details in the hard case, resulting in an incorrect VQA result. (b) With \sys{}, FastV is assigned an image-specific budget based on the corresponding energy spectrum. For the hard case, the budget is increased to 259 tokens, enabling the model to preserve critical information (e.g., labels on bottles and signs) and produce the correct answer. For the simple case, fewer tokens are allocated to the uninformative background while the answer remains correct. Green ellipses highlight regions exhibiting visible differences in token allocation, while green and red texts denote correct and incorrect answers, respectively. }\vspace{-0.1in}
    \label{fig:compare_example}
\end{figure*}

\subsubsection{Energy Spectrum as Information Richness Indicator.}
We build upon prior studies that employ Singular Value Decomposition (SVD) to compactly represent image and video signals \cite{svd1,svd2} and characterize feature matrices through their singular-value spectra \cite{svd3,svd4,svd5}. Inspired by these findings, we leverage the spectrum of the visual feature matrix to quantify the concentration of image information. Specifically, we decompose the visual feature matrix using SVD as follows:
\begin{equation}
(\mathbf{U}, \mathbf{S}, \mathbf{V}^\top) = \text{SVD}(\mathbf{Z}^V),
\end{equation}
where the singular values $\sigma = \text{diag}(\mathbf{S})$ characterize the energy captured along the corresponding singular directions. 
Accordingly, each squared singular value \(\sigma_i^2\) represents the spectral energy captured by the \(i\)-th singular direction, and the total energy of the feature space is given by their sum: 
\begin{equation}
\text{E}_{\text{total}}=\sum_{i=1}^{n} \sigma_i^2,
\end{equation}
where \(n = \min(n_v, d_v)\) denotes the total number of singular values. This energy decomposition enables us to characterize the information density of an image in terms of spectral concentration. Highly redundant images, e.g., simple case in Figure~\ref{fig:compare_example} exhibit a rapidly decaying spectrum, with only a few principal components accounting for most of the total energy. 
In contrast, complex scenes, e.g., hard case in Figure~\ref{fig:compare_example} containing densely distributed information produce a flatter spectrum, suggesting that their information is spread across a broader range of components. 

\subsubsection{Determine Image-specific Token Budget.}
Based on the energy spectrum of each image, we derive an image-specific token budget by identifying the smallest rank \(k^*\) whose cumulative energy accounts for at least a predefined proportion \(\tau\) of the total spectral energy: 
\begin{equation}
    k^*=\min\left\{k\in\{1,\dots,n\}\mid\frac{\sum_{i=1}^{k}\sigma_{i}^{2}}{\sum_{i=1}^{n}\sigma_{i}^{2}}\ge\tau\right\}
    \label{eq:rank_budget}
\end{equation}
The complete procedure is summarized in Algorithm~\ref{alg:energy_rank}. 

\noindent\textbf{Advantages of Controlling Token Budget from an Energy Perspective.}
Compared with directly specifying a fixed token budget \(k\), controlling the image-specific budget \(k^*\) through energy ratio \(\tau\) offers two primary advantages. 

\noindent\textbf{First,} it enables the token budget to be adaptively determined for each image based on its information richness. As illustrated by two example images in Figure~\ref{fig:compare_example}, their energy spectra, shown in the rightmost plots, exhibit different spectral decay patterns. Thus, under the same energy ratio of \(\tau=99\%\), \sys{} naturally assigns distinct image-specific token budgets: \(k^*=259\) for the hard case and \(k^*=95\) for the simple. In contrast, assigning a uniform fixed budget, as illustrated in Figure~\ref{fig:compare_example}(a), results in an inefficient allocation of visual tokens that fails to accommodate the distinct requirements of both cases. 

\begin{figure}[!tb]
    \centering
    \includegraphics[width=0.8\linewidth]{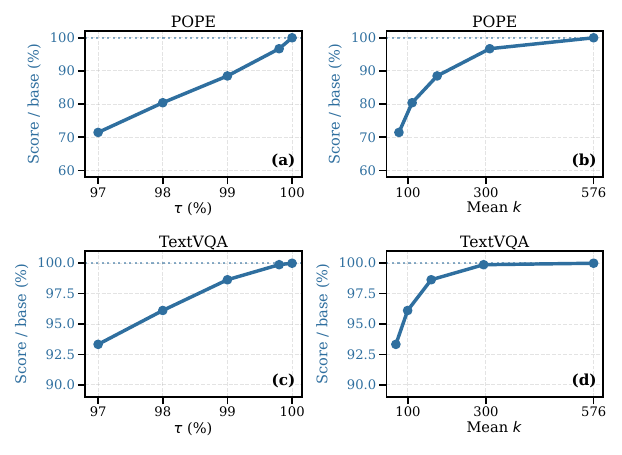}\vspace{-0.15in}
    \caption{Energy ratio \(\tau\) v.s. token count \(k\) as an operating point for task performance and token efficiency trade-off. (a)(c) We vary the energy ratio \(\tau\) from \(97.0\%\) to \(100\%\) for FastV+\sys{} and evaluate its performance on POPE/TextVQA benchmarks. (b)(d) For each setting in (a)(c), we examine the token count \(k\). Compared with token count \(k\), energy ratio \(\tau\) provides a smoother operating point.}\vspace{-0.1in}
    \label{fig:tau_sensitivity}
\end{figure}

\noindent\textbf{Second,} the energy ratio \(\tau\) provides a smooth operating point for balancing accuracy and token compression. As shown in Figure~\ref{fig:tau_sensitivity}, directly varying the token budget \(k\) leads to a highly nonlinear performance trajectory, with rapid changes followed by gradual saturation. By contrast, adjusting \(\tau\) produces a substantially smoother, approximately linear variation in performance. We do not claim that a universally optimal \(\tau\) exists across all deployment scenarios. Rather, \(\tau\) facilitates adaptation to practical requirements: a lower \(\tau\) may be preferred when memory constraints are the dominant factor and the application is not safety-critical. To ensure fair comparisons, we use the same prespecified \(\tau\) values across all corresponding experimental settings.

\section{Evaluation}
\label{sec:exp}

\begin{figure*}[t]
    \centering
    \includegraphics[width=\textwidth]{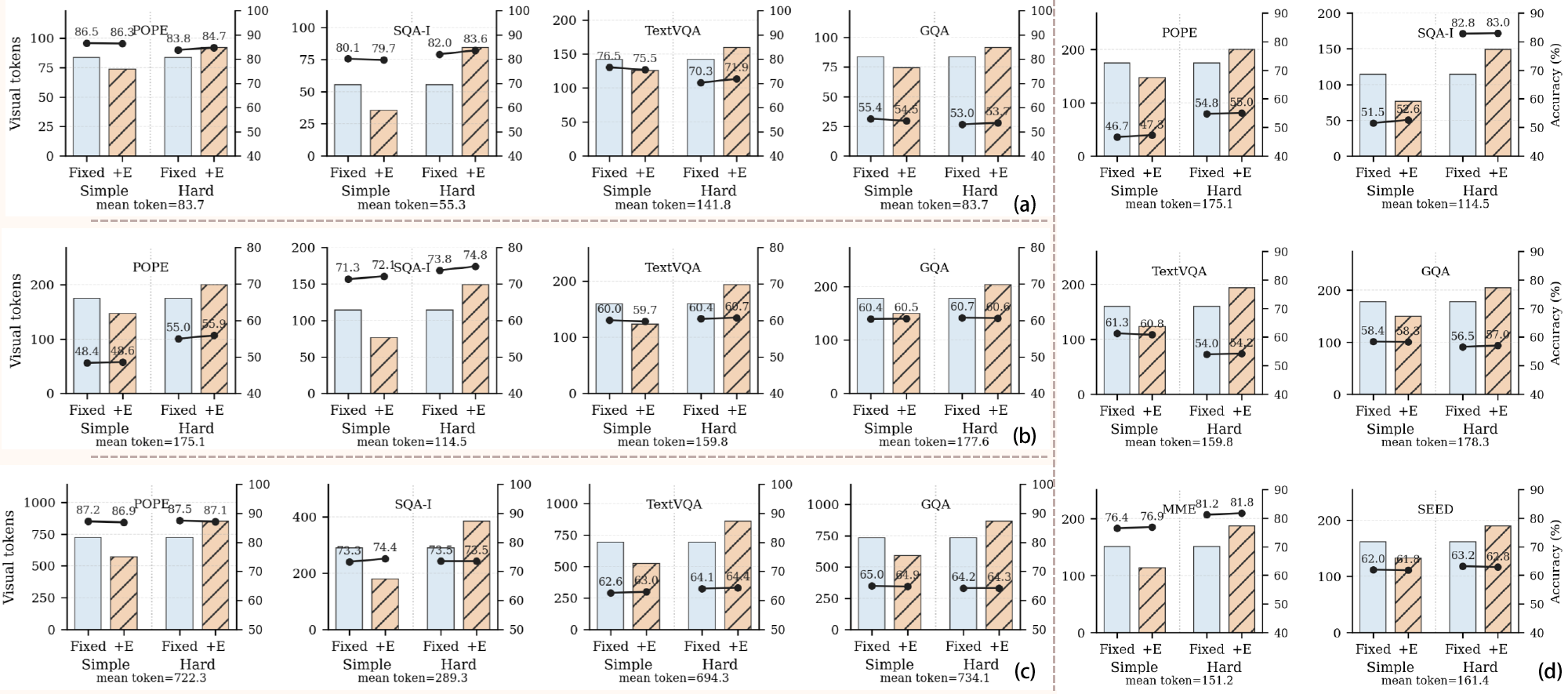}\vspace{-0.1in}
    \caption{Matched budget results with FastV on (a) Qwen2.5-VL-3B at $\tau=95.0\%$, (b) LLaVA-1.5-13B at $\tau=99.0\%$, (c) LLaVA-NeXT-8B at $\tau=99.0\%$, and (d) LLaVA-1.5-7B at $\tau=99.0\%$. Bars show retained visual tokens, and lines show subset accuracy. Simple and hard denote samples assigned fewer and more tokens than the dataset mean, respectively.}\vspace{-0.1in}
    \label{fig:results}
\end{figure*}

In this section, we evaluate \sys{} by addressing the following research questions: 

\noindent\(\bullet\) \textbf{RQ1}: Can E-AdaPrune adaptively assign token budgets to images with varying levels of information richness, reducing token usage while preserving VLM performance on simple cases and allocating more tokens to hard cases to improve performance, and generalize across VLM backbones?

\noindent\(\bullet\) \textbf{RQ2}: Can E-AdaPrune be seamlessly integrated with diverse token selectors employing different pruning criteria and consistently deliver performance improvements?

\noindent\(\bullet\) \textbf{RQ3}: Does E-AdaPrune introduce an acceptable level of computational overhead for token budget determination?
\subsection{Experimental Setup}
\subsubsection{Evaluation Benchmarks.}
We adopt comprehensive VLM benchmarks for evaluation, including POPE \cite{pope}, SQA \cite{sqa}, TextVQA \cite{textvqa}, GQA \cite{gqa}, SEED-Bench (SEED) \cite{seed}, and MME \cite{mme}, which provide sample-level performance measurements for fine-grained analysis. We further report overall results on MMBench/MMBench$^{\text{CN}}$ \cite{mmbench}, and MM-Vet \cite{mmvet}.

\subsubsection{VLM Backbones.}
We evaluate our method across VLM backbones from diverse model families and at different scales, including Qwen2.5-VL-3B \cite{qwen2.5vl}, LLaVA-1.5-7B,  LLaVA-NeXT-8B \cite{llavanext}, and LLaVA-1.5-13B \cite{llava} to address RQ1. 

\subsubsection{Token Selectors.}
We integrate \sys{} with diverse token selection strategies, including FastV \cite{fastv}, PyramidDrop (PDrop) \cite{pdrop}, and VisionZip \cite{visionzip}, to evaluate its effectiveness in a plug-and-play setting (RQ2). Their vanilla implementations employ fixed token budgets. Detailed descriptions of these selectors are provided in the Appendix. 

\subsection{Evaluation across Various VLM backbones (RQ1)}

\subsubsection{Evaluation Methodology.}
For each benchmark, we apply \sys{} with a predefined energy ratio \(\tau\) to determine an image-specific token budget. 
We then compute the average token budget across the benchmark and use this value as the uniform budget for the fixed-budget baseline, thereby ensuring a fair comparison under the same average token consumption. 
All methods use deterministic decoding and the same evaluation scripts and answer-parsing rules.
Full implementation details and reproducibility settings are provided in the Appendix.
For fine-grained analysis, we partition the samples into two groups: \emph{simple cases}, which receive fewer tokens under \sys{} than under the fixed-budget baseline, and \emph{hard cases}, which receive more tokens. 
These are operational labels based on the assigned rank, not independent annotations of semantic difficulty. Using FastV as the token selector, we evaluate the two groups separately across diverse VLM backbones and report the results in Figure~\ref{fig:results}. 
Due to space constraints, we present results for one representative value of \(\tau\) in the main text and provide results for additional settings in the Appendix. 

\begin{figure*}[t]
    \centering
    \includegraphics[width=\linewidth]{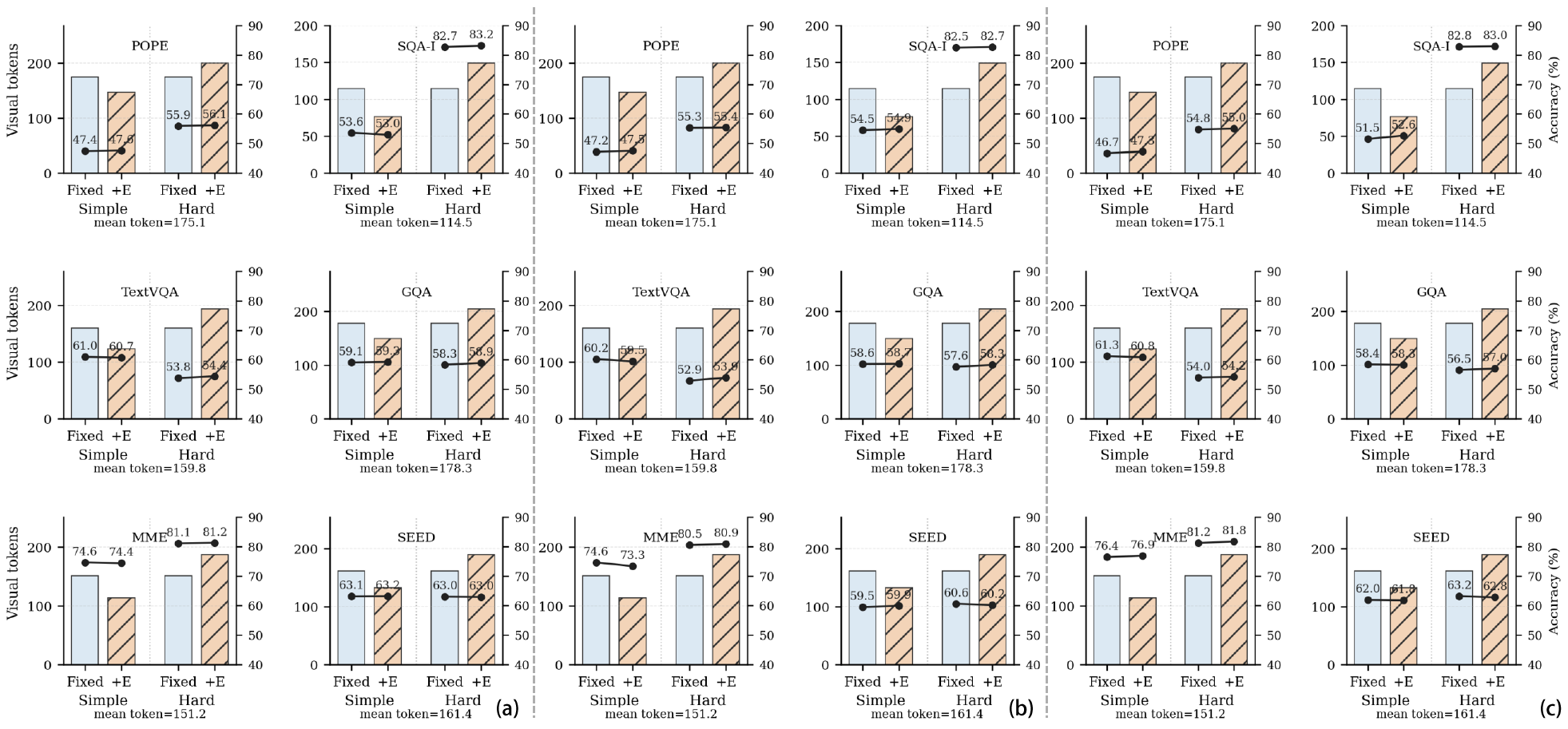}\vspace{-0.1in}
    \caption{Matched budget results on LLaVA-1.5-7B at $\tau=99.0\%$ with (a) VisionZip, (b) PDrop, and (c) FastV. Bars show retained visual tokens, and lines show subset accuracy.}\vspace{-0.1in}
    \label{fig:selectors_99}
\end{figure*}

\subsubsection{Main Results.}
For each model and benchmark, we compare adaptive allocation with a fixed FastV baseline under the same average visual token budget, and separately analyze simple and hard cases. We evaluate representative accuracy-efficiency operating points using $\tau=95.0\%$ for Qwen2.5-VL-3B and $\tau=99.0\%$ for the LLaVA backbones. 
Figure~\ref{fig:results} shows that across all four backbones, \sys{} removes a substantial amount of redundant computation from simple cases and converts the saved budget into larger gains on information rich cases.
Notably, on SQA$^\mathrm{I}$ with LLaVA-1.5-13B, \sys{} reduces the average token count for simple cases from 114.5 to 76.5, saving 33.2\% of the tokens while improving accuracy by 1.16\% from 71.3 to 72.1. The saved budget is redirected to hard cases, where the average token count increases by 30.4\% from 114.5 to 149.4 and accuracy improves by 1.42\% from 73.8 to 74.8. An even larger hard case gain appears on Qwen2.5-VL-3B. On SQA$^\mathrm{I}$, simple cases use 35.8\% fewer tokens with a relative accuracy decrease of only 0.52\%, while hard cases receive 52.5\% more tokens and achieve a 1.94\% relative accuracy improvement from 82.0 to 83.6. 
The same favorable allocation pattern appears on LLaVA-NeXT-8B and LLaVA-1.5-7B. On TextVQA with LLaVA-NeXT-8B and MME with LLaVA-1.5-7B, \sys{} reduces the simple case token budget by 24.6\% while improving accuracy by 0.63\% and 0.67\%, respectively. It then increases the hard case token budget by approximately 24\%, yielding further accuracy gains of 0.37\% and 0.72\%, respectively. Additional energy ratios in the Appendix reproduce this allocation pattern and confirm that \sys{} remains effective across various $\tau$ settings.

\subsection{Evaluation across Various Token Selectors (RQ2)}

\subsubsection{Evaluation Methodology.}
We follow the methodology described in the previous section, conducting a fine-grained analysis of simple and hard cases. Here, we use LLaVA-1.5-7B as the VLM backbone and report the results across diverse token selectors in Figure~\ref{fig:selectors_99} and Table \ref{tab:remain_results}.

\begin{table}[!tb]
\centering
\small
\renewcommand{\arraystretch}{1.2}
\resizebox{1\linewidth}{!}{
\begin{tabular}{ccccc}
\toprule
Method & MMVet & MMB & MMB$^{\text{CN}}$ & Avg. \\
\midrule
\textbf{Base} & 31.1 & 64.3 & 58.3 & 100.0\% \\
\midrule
\rowcolor{colorRatio} $\tau$=99.8\% & 298 & 261 & 261 & 273 \\
\midrule
\multirow{1}{*}{FastV} & 28.9(92.9\%)& 64.0(99.5\%) & 58.3(100.0\%) & \multirow{1}{*}{97.5\%} \\
\rowcolor{colorE} \cellcolor{colorE} FastV+E & 30.0(96.5\%) & 64.7(100.6\%) & 58.7(100.7\%) & \cellcolor{colorE} 99.3\% \\
\midrule
\multirow{1}{*}{PDrop} & 31.9(102.6\%) & 65.0(101.1\%) & 58.5(100.3\%) & \multirow{1}{*}{101.3\%} \\
\rowcolor{colorE} \cellcolor{colorE} PDrop+E & 33.5(107.7\%) & 65.2(101.4\%) & 59.1(101.4\%) & \cellcolor{colorE} 103.5\% \\
\midrule
\multirow{1}{*}{VisionZip} & 31.1(100.0\%) & 64.0(99.5\%) & 57.4(98.5\%) & \multirow{1}{*}{99.3\%} \\
\rowcolor{colorE} \cellcolor{colorE} VisionZip+E & 31.9(102.6\%) & 64.1(99.7\%) & 58.6(100.5\%) & \cellcolor{colorE} 100.9\% \\
\bottomrule
\end{tabular}
}
\caption{Overall LLaVA-1.5-7B results at $\tau=99.8\%$ on MMVet, MMB, and MMB$^{\text{CN}}$. "+E" denotes energy based adaptive budgeting. Avg. is computed over the three reported relative scores.}
\label{tab:remain_results}
\vspace{-0.1in}
\end{table}

\begin{table}[!tb]
\centering
\small
\renewcommand{\arraystretch}{1.1}
\resizebox{0.85\linewidth}{!}{
\begin{tabular}{llccc}
\toprule
\multirow{2}{*}{Method} & \multirow{2}{*}{Metric} & \multicolumn{3}{c}{Compression Ratio} \\
\cmidrule(lr){3-5}
& & 0\% & 55\% & 78\% \\
\midrule
\multirow{2}{*}{VisionZip (Fixed Token)} & MMB & 64.3 & 64.0 & 62.2 \\
& Time (s) & 933 & 722 & 628 \\
\midrule
\rowcolor{colorE} \cellcolor{colorE} & MMB & \textbf{64.3} & \textbf{64.1} & \textbf{62.5} \\
\rowcolor{colorE}\multirow{-2}{*}{VisionZip+\sys{}} & Time (s) & 933 & 843 & 781 \\
\bottomrule
\end{tabular}
}
\caption{Efficiency and Performance with Conventional SVD.}
\label{tab:efficiency_comparison}
\vspace{-0.1in}
\end{table}

\subsubsection{Main Results.}
We evaluate the generalization of \sys{} across VisionZip, PDrop, and FastV on LLaVA-1.5-7B. For each selector, we retain its original token scoring rule and compare adaptive allocation with its fixed budget counterpart under the same average visual token budget. Figure~\ref{fig:selectors_99} shows that \sys{} consistently reduces tokens for simple cases and redirects the saved budget to hard cases across all three selectors. Most notably, with FastV on MME, simple cases use 24.6\% fewer tokens while accuracy improves by 0.67\%, whereas hard cases receive approximately 24\% more tokens and improve by 0.72\%. VisionZip on POPE shows the same favorable allocation, with 16.1\% fewer tokens and a 0.45\% accuracy improvement for simple cases, and 14.5\% more tokens and a 0.45\% improvement for hard cases. With PDrop on GQA, the corresponding token changes are 16.0\% fewer and 15.0\% more, while accuracy improves by 0.11\% and 1.07\%. These consistent gains show that \sys{} generalizes across selectors with different pruning criteria without changing their selection mechanisms.

To complement the fine-grained results at $\tau=99.0\%$, we evaluate a higher-retention operating point at $\tau=99.8\%$ on three additional benchmarks. Since sample level correctness is unavailable for MMVet, MMBench, and MMBench$^{\text{CN}}$, we report their overall scores in Table~\ref{tab:remain_results}. \sys{} consistently improves all three selectors across the three benchmarks, increasing their average relative scores by up to 2.17\%. The complete results in Table~\ref{tab:remain_results} further show improvements in all nine selector-benchmark combinations.

\subsection{Efficiency Evaluation (RQ3)}
\subsubsection{Evaluation Methodology.} 
Compared with a fixed visual token budget, \sys{} performs SVD on the visual feature matrix to derive an adaptive and image-specific token budget. 
To quantify the resulting computational overhead, we use VisionZip as the token selector because it physically removes redundant tokens and thus provides a faithful measurement of computational efficiency. By contrast, FastV and PDrop primarily mask token computation without reducing the actual token sequence, potentially biasing efficiency measurements. 
We compare the execution times of VisionZip with a fixed token budget and VisionZip+\sys{} on MMBench, using LLaVA-1.5-7B as the VLM backbone and an NVIDIA A6000 GPU. 
Following previous evaluation methodology, we ensure that both have the same average token consumption.
The results are in Table~\ref{tab:efficiency_comparison}. 

\subsubsection{Main Results.} Although the total execution time decreases as the compression ratio increases, VisionZip+\sys{} still incurs a noticeably longer runtime than VisionZip alone. At \(78\%\) compression ratio, VisionZip requires \(628\)s, whereas VisionZip+\sys{} requires \(781\)s. 
Since both methods operate under the same average token consumption, this additional latency does not arise from VLM inference but from the budget determination procedure in \sys{}. In particular, performing a full SVD on the visual feature matrix introduces non-negligible computational overhead. 

\subsubsection{Improving Efficiency via rSVD.}
We then investigate whether standard randomized SVD (rSVD) \cite{rSVD,rsvd2} can alleviate this computational overhead. rSVD first projects the visual features onto a lower-dimensional random subspace and subsequently estimates the dominant singular-value spectrum from the compressed matrix. A complete description of the algorithm is provided in the supplementary material. 
Table~\ref{tab:rsvd_ablation} demonstrates that rSVD substantially alleviates the computational bottleneck. Exact SVD incurs an additional latency of \(35\) ms per image, whereas rSVD with a target dimension of \(t=300\) and \(q=2\) power iterations reduces this overhead to \(8\) ms while preserving the same MMB score of \(62.5\). 
Regarding the hyperparameter \(q\) in rSVD, omitting it (\(q=0\)) can lead to an overestimation of the average visual-token budget, particularly at smaller target dimensions. In practice, \(q=2\) provides a stable approximation to exact SVD, while larger values of \(q\) yield negligible benefits relative to their latency cost.

\begin{table}[!tb]
\centering
\small
\renewcommand{\arraystretch}{1.1}
\resizebox{1\linewidth}{!}{
\begin{tabular}{lccccc}
\toprule
Method & \makecell{Dim($t$)} & \makecell{Avg. Token} & \makecell{Time($s$)} & \makecell{Latency($ms$)} & MMB \\
\midrule
\textbf{Base} & - & 576 & 895 & - & 64.3 \\
\midrule
\textbf{VisionZip} & - & 130 & 628 & - & 62.2 \\
\midrule
\textbf{basic SVD} & 1024 & 130 & 781 & 35 & 62.5 \\
\midrule
rSVD ($q=2$) & 500 & 130 & 701 & 17 & 62.5 \\
rSVD ($q=1$) & 500 & 130 & 687 & 13 & 62.5 \\
rSVD ($q=0$) & 500 & 131 & 695 & 15 & 62.4 \\
\midrule
rSVD ($q=2$) & 400 & 130 & 702 & 17 & 62.5 \\
rSVD ($q=1$) & 400 & 130 & 678 & 11 & 62.5 \\
rSVD ($q=0$) & 400 & 136 & 685 & 13 & 62.8 \\
\midrule
\rowcolor{highlightcolor} rSVD ($q=2$) & 300 & 130 & 660 & 8 & 62.5 \\
rSVD ($q=1$) & 300 & 131 & 679 & 12 & 62.4 \\
rSVD ($q=0$) & 300 & 156 & 658 & 7 & 62.8 \\
\bottomrule
\end{tabular}\vspace{-0.1in}
}
\caption{Efficiency improvement with rSVD. Latency denotes the extra per-image overhead relative to VisionZip.}
\label{tab:rsvd_ablation}
\vspace{-0.1in}
\end{table}

\subsubsection{Efficiency under Batch Serving.}

VLM deployments commonly use batched inference. Adaptive budgets improve token allocation under the same total budget. However, mixing simple and hard cases produces different sequence lengths and increases padding overhead during prefill.
We evaluate it using LLaVA-1.5-13B with VisionZip+\sys{}. We sample 1024 VQA requests from POPE/TextVQA, precompute each retained token length, and form batches of size 4, 8, and 16. We evaluate various strategies, including first-in first-out (FIFO), random shuffling, and bucketed batching that groups requests by their final prefill lengths.
We measure Time to first token (TTFT) that includes preprocessing and batched VLM prefill until the first token logits are produced, and excludes network and queueing latency. 
We also measure padding overhead as \(\textit{batch padded tokens}/\textit{true prefill tokens}-1\). We compute it over the complete prefill sequence, including retained visual tokens and text prompt tokens.
As shown in Table~\ref{tab:dynamic_batching_overhead}, unbatched serving has an average latency of \textit{170.1 ms} per request and no padding. Batched serving initially lowers latency through better GPU utilization and lower kernel launch overhead. Padding increases with batch size because larger batches are more likely to contain requests with long token budgets. Bucketed batching reduces both TTFT and padding by grouping requests with similar lengths. Thus, adaptive visual token budgets remain efficient in batched serving, while length aware scheduling limits their padding cost.

\begin{table}[!tb]
\centering
\small
\renewcommand{\arraystretch}{1.1}
\resizebox{0.85\linewidth}{!}{
\setlength{\tabcolsep}{3.0pt}
\begin{tabular}{llcccc}
\toprule
Metric & Schedule & B1 & B4 & B8 & B16 \\
\midrule
\multirow{3}{*}{TTFT\textit{(ms/request)}} & FIFO & \multirow{3}{*}{170.1} &  164.5 & 170.7 & 178.1\\
 & Random & & 164.8 & 171.1 & 178.5\\
 & Bucketed & & \textbf{159.3} & \textbf{164.7} & \textbf{170.3}\\
\midrule
\multirow{3}{*}{Pad(\%)} & FIFO & \multirow{3}{*}{0} &  18.0 & 28.4 & 40.2\\
 & Random & & 18.6 & 28.9 & 40.8\\
 & Bucketed & & \textbf{11.8} & \textbf{20.8} & \textbf{30.8}\\
\bottomrule
\end{tabular}
}
\caption{Efficiency under Batch Serving.}
\label{tab:dynamic_batching_overhead}
\vspace{-0.1in}
\end{table}

\section{Conclusion}
\label{sec:conclusion}
This paper identifies fixed visual token budgeting as an independent bottleneck in efficient VLM inference. 
We introduced \sys{}, a training-free adaptive budgeter that estimates image-specific visual token budgets from spectral energy and passes them to existing token selectors. 
Under matched average token budgets, \sys{} successfully adapts the token budget to images with varying information richness, improving inference efficiency or task accuracy across diverse VLMs, token selectors, and benchmarks.


\newpage
\bibliography{aaai2027}

\newpage
\appendix
\section{Appendix}
\subsection{Compared Baselines}

We integrate \sys{} with three representative visual token pruning strategies. FastV \cite{fastv} is a plug-and-play inference acceleration method that identifies inefficient attention in deep layers and prunes low-attention visual tokens after a specific layer. PyramidDrop (PDrop) \cite{pdrop} progressively reduces visual redundancy by dropping image tokens across multiple stages, maintaining full tokens in shallow layers while increasing sparsity in deeper layers. VisionZip \cite{visionzip} is a text-agnostic pruning approach that selects dominant informative tokens and merges the remaining contextual tokens based on semantic similarity before feeding them into the LLM.

\subsection{Implementation Details}
For matched-budget comparisons, we first apply a fixed energy ratio, e.g., $\tau=99.8\%$ or $\tau=99.0\%$, to each benchmark image and compute its adaptive token count. The dataset-level mean count is then used as the fixed budget $k$ for the corresponding static baseline, so adaptive and static variants use the same average visual-token budget.

We keep the default baseline configurations except for budget selection. FastV \cite{fastv} prunes visual tokens after layer 2. PDrop \cite{pdrop} uses pruning layers [8, 16, 24] and clamps $\rho$ to [0.4, 0.6] for $\tau=99.8\%$ and [0.6, 0.8] for $\tau=99.0\%$. VisionZip \cite{visionzip} uses 30 contextual tokens. Qwen2.5-VL follows the same matched-budget protocol with FastV. Latency measurements use one NVIDIA RTX A6000 with PyTorch 2.1.2, CUDA 12.1, and driver 550.67.

\subsection{Randomized SVD Algorithm}
Despite the adaptive benefits of spectral analysis, performing a full SVD on the visual feature matrix $\mathbf{Z}^V \in \mathbb{R}^{n_v \times d_v}$ introduces significant computational overhead. Specifically, exact SVD requires $O(n_v d_v \min(n_v, d_v))$ operations, which may offset the latency gains achieved through token pruning. To mitigate this, rSVD \cite{rSVD,rsvd2} first projects $\mathbf{Z}^V$ onto a smaller random subspace to identify its most significant range, and then performs decomposition on this compressed representation. By using a target dimension $t$, the dominant singular spectrum can be approximated with complexity $O(n_v d_v t + t^2 d_v)$.

\begin{algorithm}[H]
\caption{Efficient Energy-based Rank Selection via Randomized SVD}
\label{alg:rsvd_energy_rank}
\textbf{Input}: $\mathbf{Z}^V$, energy ratio $\tau$, oversampling $p$, power iter $Q$, range $[k_{min}, k_{max}]$ \\
\textbf{Output}: $k^*$
\begin{algorithmic}[1]
\STATE $E_{total} \leftarrow \|\mathbf{Z}^V\|_F^2$

\STATE $k_{sub} \leftarrow \min(k_{max} + p, d_v)$
\STATE $\Omega \leftarrow \text{randn}(d_v, k_{sub})$
\STATE $Y \leftarrow \mathbf{Z}^V \Omega$

\FOR{$q = 1$ to $Q$}
    \STATE $Y \leftarrow \mathbf{Z}^V (\mathbf{Z}^{V\top} Y)$
    \STATE $Y, \_ \leftarrow QR(Y)$
\ENDFOR

\STATE $B \leftarrow Y^\top \mathbf{Z}^V$
\STATE $\tilde{\sigma} \leftarrow \text{SingularValues}(B)$

\STATE $k^* \leftarrow \min \{ k \mid \frac{\sum_{i=1}^k \tilde{\sigma}_i^2}{E_{total}} \ge \tau \}$

\STATE \textbf{return} $k^*$

\end{algorithmic}
\end{algorithm}

\subsection{Additional Results}
We present the additional results here.
\begin{figure*}[tb]
    \centering
    \includegraphics[width=0.95\textwidth]{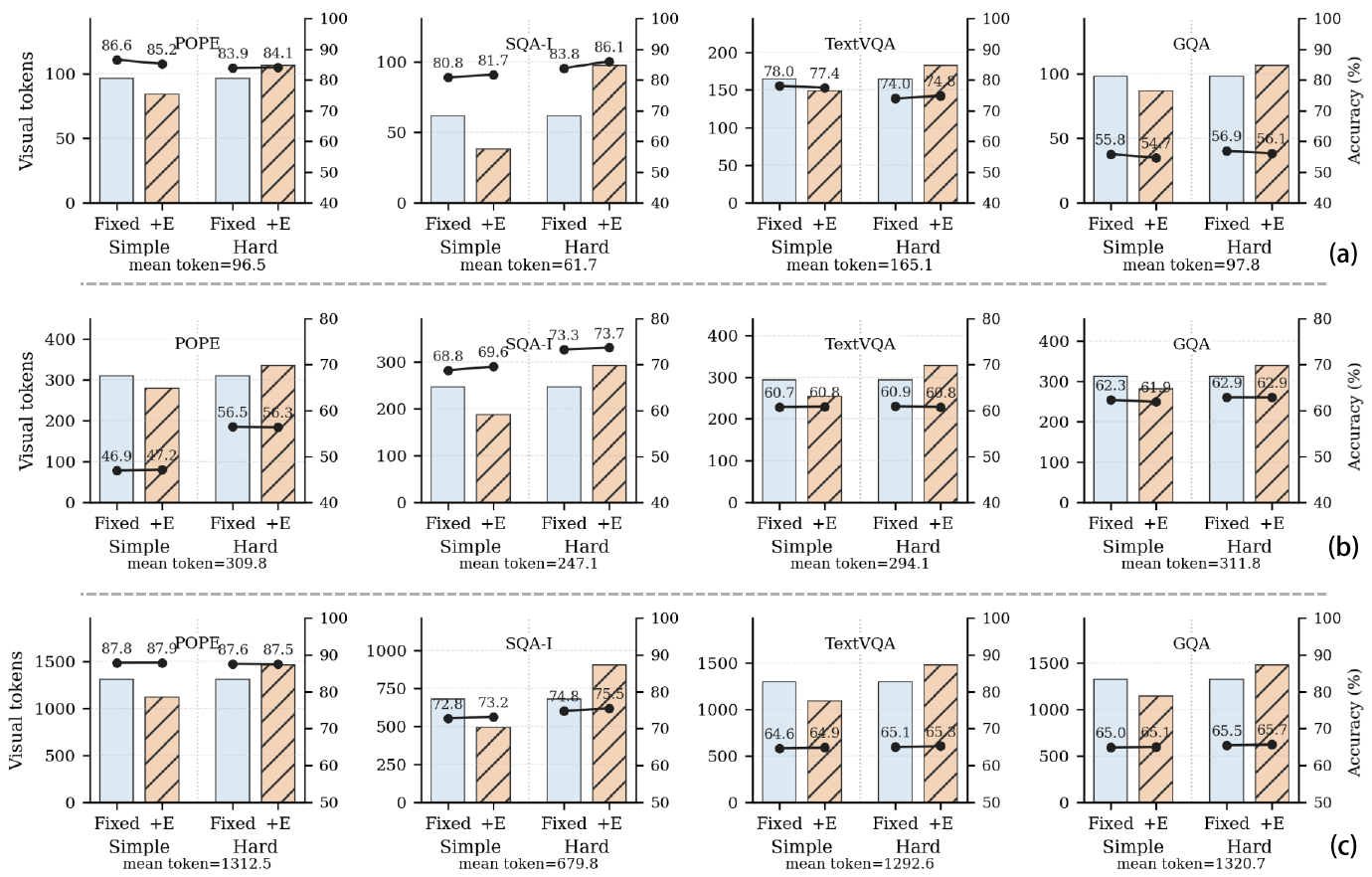}
    \caption{Matched budget results on (a)Qwen2.5-VL-3B, $\tau$=96\%, (b)LLaVA-1.5-13B, $\tau$=99.8\%, and (c)LLaVA-NEXT-8B, $\tau$=99.8\%.}
    \label{fig:tau=998}
\end{figure*}
\begin{figure*}[tb]
    \centering
    \includegraphics[width=\textwidth]{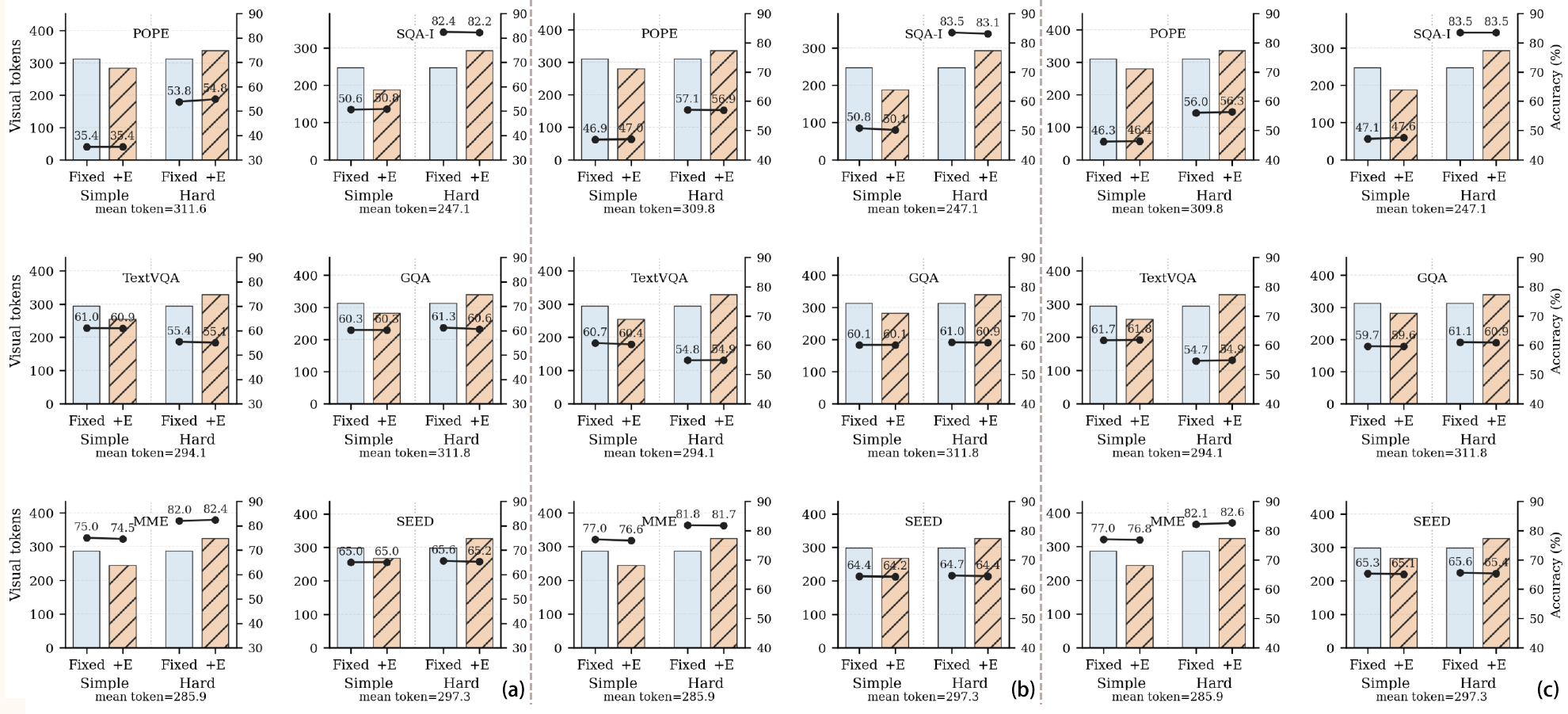}
    \caption{Matched budget results on LLaVA-1.5-7B with (a) VisionZip, (b) PDrop, and (c) FastV, all using $\tau=99.8\%$.}
    \label{fig:selectors_998}
\end{figure*}


\end{document}